%% file: main.tex
\documentclass[10pt]{article} 
\usepackage[preprint]{rlc}

\usepackage{amssymb}      
\usepackage{mathtools}          
\usepackage{mathrsfs}     
\numberwithin{equation}{section}
\usepackage{graphicx}           
\usepackage[space]{grffile}     
\usepackage{url}    
\usepackage{xspace}
\usepackage{subfigure}
\usepackage{tikz,ifthen}
\usepackage{wrapfig}

\input{imports/drawing}

\title{Aquatic Navigation: A Challenging Benchmark for Deep Reinforcement Learning}

\author{Davide Corsi\\
        dcorsi@uci.edu \\
        Department of Computer Science\\
        University of California, Irvine
        \And
        Davide Camponogara  \\
        davide.camponogara@studenti.univr.it \\
        Department of Computer Science\\
        University of Verona
        \And
        Alessandro Farinelli  \\
        alessandro.farinelli@univr.it \\
        Department of Computer Science\\
        University of Verona       
}

\begin{document}

\maketitle

\begin{abstract}
An exciting and promising frontier for Deep Reinforcement Learning (DRL) is its application to real-world robotic systems. While modern DRL approaches achieved remarkable successes in many robotic scenarios (including mobile robotics, surgical assistance, and autonomous driving) unpredictable and non-stationary environments can pose critical challenges to such methods.  These features can significantly undermine fundamental requirements for a successful training process, such as the Markovian properties of the transition model. To address this challenge, we propose a new benchmarking environment for aquatic navigation using recent advances in the integration between game engines and DRL. In more detail, we show that our benchmarking environment is problematic even for state-of-the-art DRL approaches that may struggle to generate reliable policies in terms of generalization power and safety. Specifically, we focus on PPO, one of the most widely accepted algorithms, and we propose advanced training techniques (such as curriculum learning and learnable hyperparameters). Our extensive empirical evaluation shows that a well-designed combination of these ingredients can achieve promising results. Our simulation environment and training baselines are freely available to facilitate further research on this open problem and encourage collaboration in the field.
\end{abstract}

\input{sections/introduction}
\input{sections/background}
\input{sections/robot_and_sim}
\input{sections/training}
\input{sections/validation}
\input{sections/conclusion}

\subsubsection*{Acknowledgments}
The work was carried out within the Interconnected Nord-Est Innovation Ecosystem (iNEST) and received funding from the European Union Next-GenerationEU (PIANO NAZIONALE DI RIPRESA E RESILIENZA (PNRR) – MISSIONE 4 COMPONENTE 2, INVESTIMENTO 1.5 – D.D. 1058 23/06/2022, ECS00000043). 

\bibliography{main}
\bibliographystyle{rlc}
\clearpage

\appendix 
\input{appendices/curriculum-learning}
\input{appendices/safety}

\input{appendices/reward-engineering}

\end{document}

%% file: imports/drawing.tex
\usetikzlibrary{arrows,trees,backgrounds,automata,shapes,decorations,plotmarks,fit,calc,positioning,shadows,chains}

\definecolor{color0}{RGB}{239, 48, 84} 
\definecolor{color4}{RGB}{62, 255, 139} 
\definecolor{color6}{RGB}{147, 129, 255} 
\definecolor{color7}{RGB}{147, 129, 255} 
\definecolor{nnedgecolor}{RGB}{90,90,90}

\tikzstyle{every pin edge}=[<-,shorten <=1pt]
\tikzstyle{every path}=[draw=color7!50]
\tikzstyle{neuron}=[circle,fill=black!25,minimum size=17pt,inner sep=0pt]
\tikzstyle{input neuron}=[neuron, fill=color4]
\tikzstyle{output neuron}=[neuron, fill=color0]
\tikzstyle{hidden neuron}=[neuron, fill=color6]
\tikzstyle{annot} = [text width=4em, text centered]
\tikzstyle{nnedge} = [-{stealth},shorten >=0.1cm, shorten <=0.05cm,line 
width=0.8pt,nnedgecolor]
\usetikzlibrary{calc}

\usepackage{listings, lstautogobble}
\definecolor{deepblue}{rgb}{0,0,0.5}
\definecolor{deepred}{rgb}{0.8,0,0}
\definecolor{deepgreen}{rgb}{0,0.5,0}

\lstdefinestyle{mystyle}{  
    commentstyle=\color{deepgreen},
    keywordstyle=\color{deepred},
    stringstyle=\color{deepblue},
    basicstyle=\scriptsize,
    captionpos=b,
    frame=lines
}

%% file: sections/introduction.tex
\section{Introduction}
\label{sec:introduction} 
In recent years, Deep Reinforcement Learning (DRL) methods have advanced rapidly and achieved impressive results in various domains. For instance, modern DRL algorithms, such as TD3 \citep{FuHoMe18}, SAC \citep{HaZhAb18}, PPO \citep{ShWoDh17}, or Rainbow \citep{HeMoHa18}, have demonstrated remarkable capabilities in solving highly complex problems, ranging from video games \citep{MnKaSi13} to complex decision-making tasks and robotic applications \citep{kober2013reinforcement, rolf2023review}. However, even the state-of-the-art algorithms struggle when dealing with unpredictable and non-stationary environments, where the basic Markovian properties may be violated \citep{marchesini2021benchmarking}. 

In this direction, the limited availability of challenging benchmarking environments, where even state-of-the-art algorithms fail to achieve optimal performance, makes it difficult to evaluate the impact of new DRL methods and advanced learning approaches. This is especially relevant in the field of robotics, where issues related to safe control are often impossible to separate from the hardware and therefore not readily available to the community as a benchmark \citep{aractingi2023controlling, akkaya2019solving}. Moreover, a common limitation shared by almost all the DRL algorithms lies in their data efficiency \citep{LiHuPr15, HaZhTu18}. In the context of robotics, this limitation assumes particular significance due to the challenging process of data collection. Collecting real-world data on the actual robot can be slow and dangerous, especially when factors such as human safety or the use of expensive hardware are involved. Consequently, the development of realistic simulators for the training process has emerged as a priority \citep{AtScLe20, pore2021safe, AmCoYe23}. 

\begin{figure}[b]
\centering
\vspace{-3pt}
\minipage{0.48\linewidth}
    \includegraphics[width=\linewidth]{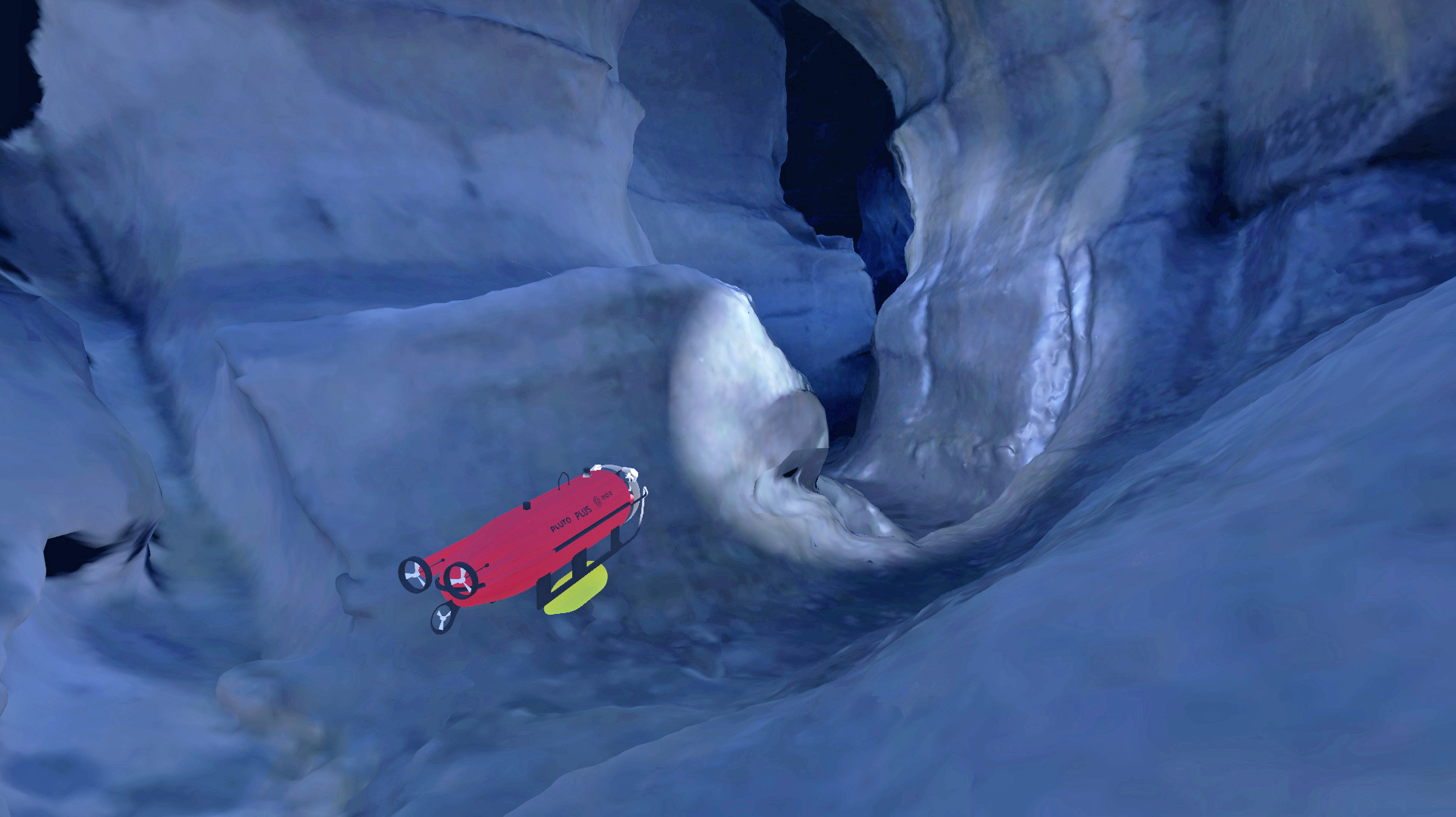}
\endminipage \hfill
\minipage{0.48\linewidth}
    \includegraphics[width=\linewidth]{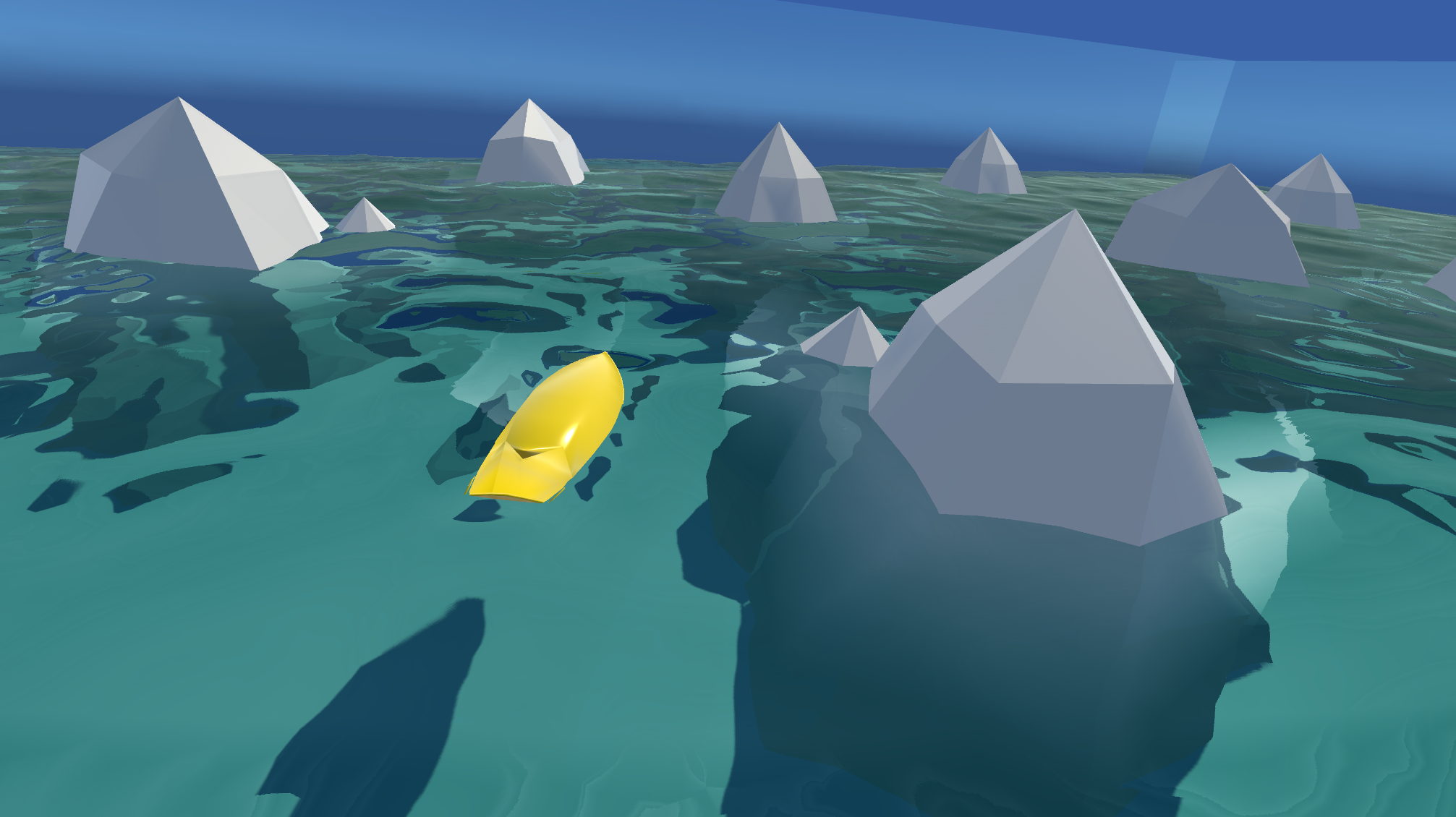}
\endminipage
\caption{The figures depict two environments within our simulator. The first figure shows our Autonomous Underwater Vehicle (AUV) navigating a 3D model of Porth yr Ogof marine cave, while the second figure shows our surface drone in one of the scenarios from our marine benchmark. Although the two environments differ in their objective, they share the main challenges introduced by the aquatic environment.}
\label{fig:intro:screenshot}
\end{figure}

Against this background, the first contribution of this paper involves developing a simulator designed for aquatic navigation, that considers both surface and underwater scenarios (see  Fig.\ref{fig:intro:screenshot}). In this type of environment, many of the aforementioned issues related to the complex and unpredictable evolution of water can arise, potentially compromising training performance. 
Specifically, our focus is on autonomous navigation, which is increasingly used for important tasks such as exploration, cable monitoring, security, and seabed mapping in oceans and lakes \citep{carreras2018sparus, wynn2014autonomous}. The simulator is tailored specifically for this purpose and addresses several critical requirements. First, it is designed to be lightweight and high-performing, enabling multiple executions to collect the substantial volume of data necessary for effective training. Second, it allows a wide range of customization possibilities to replicate different real-world environments. Finally, it strongly emphasizes realism, crucial for training autonomous agents in environments that mirror real-world challenges. Mapless navigation problems can generally be addressed with Deep Reinforcement Learning (DRL) approaches \citep{ZhMoKo17, BoDeDw16, MaFa21, MaFa22}; nevertheless, our results show that the unpredictable nature of the environment makes the task much more challenging for the DRL agent.

Our second contribution is a pipeline for training and validating a DRL agent to provide a stable baseline for comparison with future work and algorithmic improvements. We rely on Proximal Policy Optimization (PPO) \citep{ShWoDh17}, a state-of-the-art reinforcement learning algorithm that has shown groundbreaking results across a wide range of tasks. While PPO offers a general approach for reinforcement learning problems, achieving satisfactory results demands careful consideration of problem-specific configurations \citep{EnIlSa20, corsi2024analyzing} and additional optimization tricks and implementation details \citep{schulman2015high, marchesini2023improving, liang2022reducing}. Throughout this paper, we present a comprehensive set of ablation studies that support our ultimate design choices, highlighting the limitations that even state-of-the-art algorithms can have on such a complex problem. This work emphasizes the critical problem of safety in the domain of autonomous navigation. The involvement of expensive equipment and the inherent challenges associated with potential rescue operations make safety a particularly relevant concern \citep{fossen2011handbook}. Our benchmark includes additional safety requirements, making it suitable for research in the field of safe deep reinforcement learning \citep{CoMaFa21, RaAcAm19, yerushalmi2022scenario}. 

Finally, to further validate the effectiveness of our agent, we extensively test our baseline on a cave navigation scenario that is created using real-world data. In detail, we recreate the \textit{Porth yr Ogof} cave, located in South Wales; where we then deploy our trained agent without any prior knowledge of the cave's structure. The results demonstrate that the agent is able to explore the entire cave while avoiding catastrophic collisions. However, a more detailed analysis showed some limitations in terms of generalization power and safety against specific corner cases, even employing state-of-the-art solutions. For this reason, we believe that this challenging environment can be of great value to the DRL community, and not strictly limited to applications in water navigation domains. Our analysis shows that solving these tasks -- taking into account the safety and generalization aspects -- is still an open problem, and we believe that it can be considered a challenging benchmark to validate novel learning tools and algorithms. Crucially, to promote further research and collaboration in this domain, we offer open access to our simulation environment and training algorithms\footnote{\url{https://github.com/dadecampo/aquatic_navigation_envs}}\footnote{\url{https://github.com/dadecampo/SafeRLAUV}}.

%% file: sections/background.tex
\section{Related Work}
\label{sec:background}

In the previous section, we discuss a critical problem in DRL, the high amount of data necessary for an effective learning process \citep{LiHuPr15, HaZhTu18}. Collecting all these experiences can be hard in a robotic context, where expensive equipment is involved and failures can be barely tolerated. A common solution is the exploitation of realistic simulation engines. Historically, the robotic community has relied on software such as RViz (for visualization) and Gazebo (for simulation); however these solutions are not designed to support fast computation and parallel execution, both necessary requirements in a DRL context \citep{zhao2020sim, azar2023autonomous}. In contrast, standard benchmarks for DRL rely on libraries such as MuJoCo, Bullet, or PyGame to approximate the real-world dynamics, often sacrificing the accuracy of the physics simulation to obtain faster computation \citep{gronauer2022bullet, RaAcAm19}. To bridge this gap modern approaches propose the use of 3D simulation engines typically developed for video games such as Unreal \citep{de2019analysis}, Coppelia \citep{nogueira2014comparative}, or Unity3D \citep{juliani2020unity}. In this work, we focus on the latter, which has been recently successfully employed as a simulation engine for robotic research \citep{UnityRobotics}. Unity offers unique capabilities to fasten the simulation, such as a \textit{server mode} that allows computing the simulation without the rendering part, \textit{time acceleration}, and the \textit{synchronous execution} to allow easier integration with the state-of-the-art DRL libraries. In fact, a crucial advantage of Unity3D with respect to other engines is the built-in package \textit{Unity ML-Agents}, which provides full compatibility with Gym, a standardized set of API for DRL research \citep{juliani2020unity}. We believe this is a critical asset to foster the wider use of a DRL benchmark in the community as Gym is the de facto standard interface for the most popular DRL implementation (e.g., \textit{Stable-Baselines}, \textit{SpinningUP}, \textit{CleanRL}, and more). There are already underwater navigation simulators that accurately simulate the physical characteristics of these scenarios \citep{9976969, Cielak2019StonefishAA}, but they are not simulators designed for DRL, but rather for data collection and dataset generation from simulations. This represents a significant limitation for DRL practitioners, as it does not allow for a straightforward integration of the learning algorithm. For example, DRL algorithms are designed to solve variations of a Markov Decision Process, which requires a discretization of the time. This is particularly challenging in the context of complex physics simulations (e.g., water). Moreover, our environment allows for easy access to the reward (and cost) function and a clear and explicit definition of the state and action spaces. Finally, the results presented in this paper provide a fair baseline for future algorithm and approach development.

\subsection*{Navigation and Mapless Navigation}
We focus on the problem of navigating a robot through an environment, to reach a specific target position. Typically, the agent should adhere to additional constraints, that may include finding the shortest path, avoiding obstacles, or optimizing energy consumption. In the last years, this problem has gathered increasing attention, particularly due to its relevance in the context of autonomous vehicles \citep{PaYoWa17}, and it is today considered one of the classical problems in robotics. Robotic Navigation has been extensively studied over the years, resulting in various algorithmic solutions such as planning and search-based approaches \citep{lavalle2006planning, latombe2012robot}. Nevertheless, a variant of robotic navigation, known as \textit{mapless navigation}, has recently emerged as a popular problem, and a standard benchmark for DRL, that presents additional unique challenges \citep{ZhMoKo17, MaFa22}. 
In mapless navigation, the robot operates within the environment without using a map, relying solely on its local observations. This configuration introduces additional complexities, as the absence of a map hinders the use of conventional planning-based methods. Moreover, limited sensor information makes the problem \textit{partially observable}, giving rise to additional challenges such as sensor noise and the uncertainty of action outcomes \citep{MaFa22}.  State-of-the-art solutions for mapless navigation suggest exploiting DRL techniques to generate policies capable of controlling the autonomous vehicle; these solutions have demonstrated exceptional performance \citep{BoDeDw16} and are nowadays considered a clear example of the DRL's potential. Moreover, recent works show that these kinds of problems can be accomplished by employing relatively simple and small DNN architectures, which is essential for enabling on-board control of the robot, where resource constraints require compact models.

%% file: sections/robot_and_sim.tex
\section{Simulation Environment}
\label{sec:robot_sim}
In this section, we introduce our environment, presenting the first contribution of the paper: a realistic underwater simulator based on the Unity3D game engine. Unity has emerged as a powerful tool for the development of Reinforcement Learning agents in simulated scenarios, especially in the domain of robotics research \citep{juliani2020unity}. As a fundamental building block for our experiments, we developed a virtual environment tailored to closely emulate the challenges of aquatic environments. To replicate the hydrodynamic aspects of water, we rely on ZibraAI Liquids \citep{ZibraHowDoesWorks}, a state-of-the-art solution for real-time 3D liquid simulation. This versatile tool provides us with the flexibility to manipulate a wide array of parameters, encompassing liquid physics settings and interaction dynamics with other physical objects in the environment. 

\begin{figure}[t]
\minipage{0.48\linewidth}
    \includegraphics[width=\linewidth]{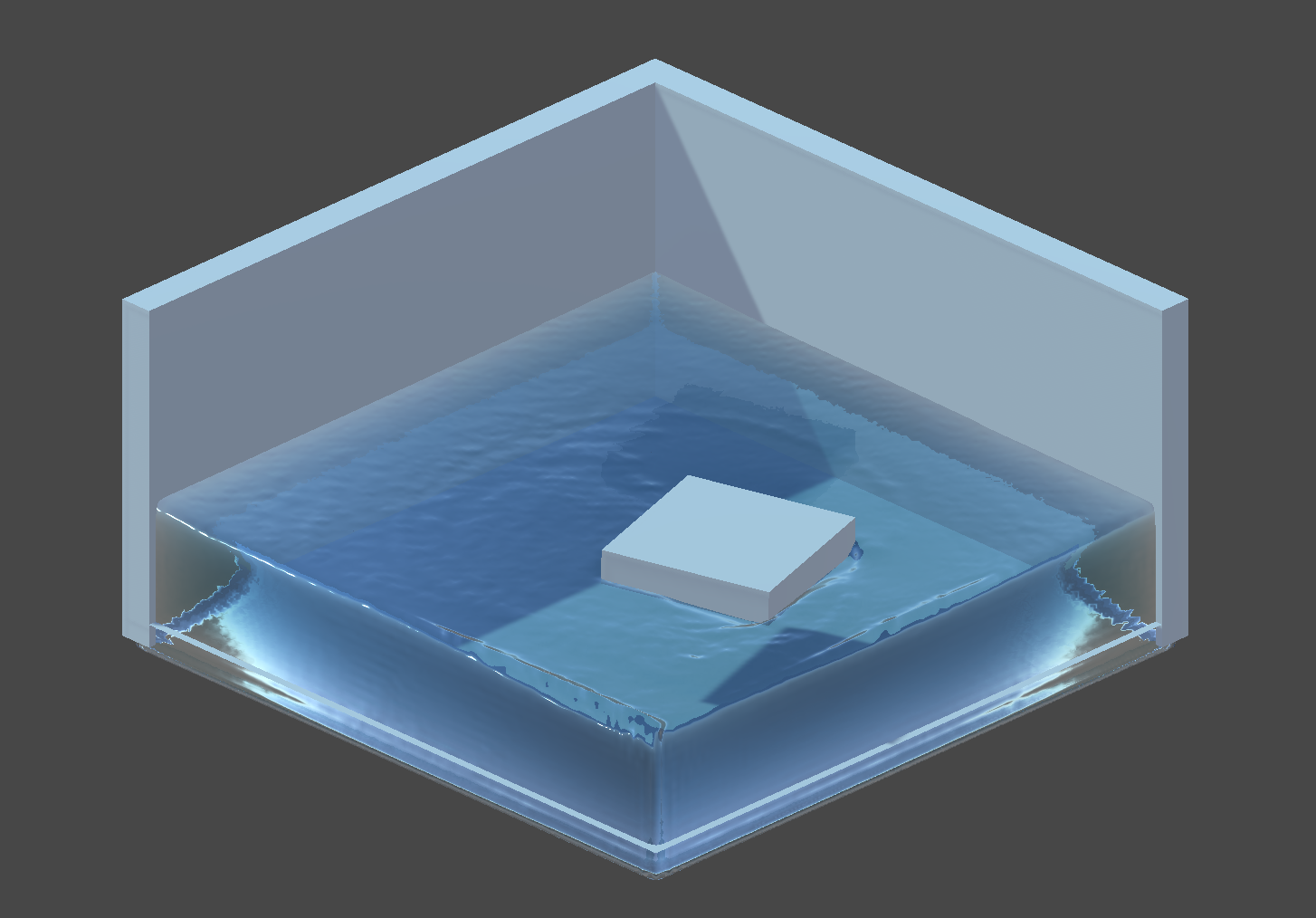}
    \caption{Viscous liquid.}
    \label{fig:robot:100_viscosity}
\endminipage \hfill
\minipage{0.495\linewidth}
    \includegraphics[width=\linewidth]{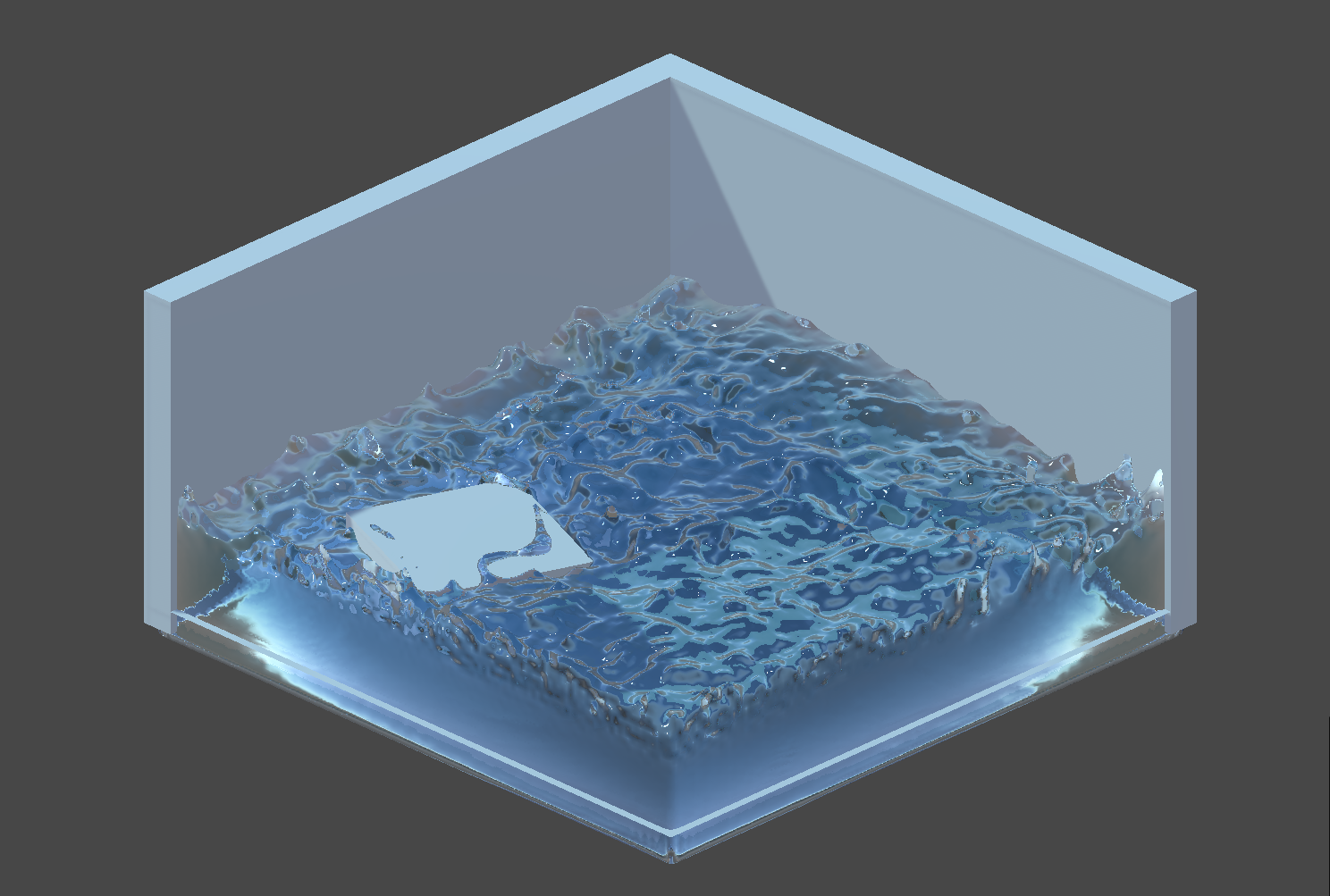}
    \caption{Runny liquid.}
    \label{fig:robot:0_viscosity}
\endminipage
\end{figure}

\subsection{Simulation of Fluid Behavior}
Simulating the behavior of fluids has always been considered a hard challenge, especially due to the multitude of intricate physical forces involved \citep{9976969, Cielak2019StonefishAA}. 
Moreover, precisely reproducing all these forces in real time remains impractical, necessitating a significant degree of approximation. ZibraAI enables the creation of water zones that can be precisely parameterized to adjust viscosity, surface tension, and other essential characteristics. 
\begin{wrapfigure}{r}{0.6\textwidth}
    \vspace{-5pt}
        \begin{center}
        \includegraphics[width=0.55\textwidth]{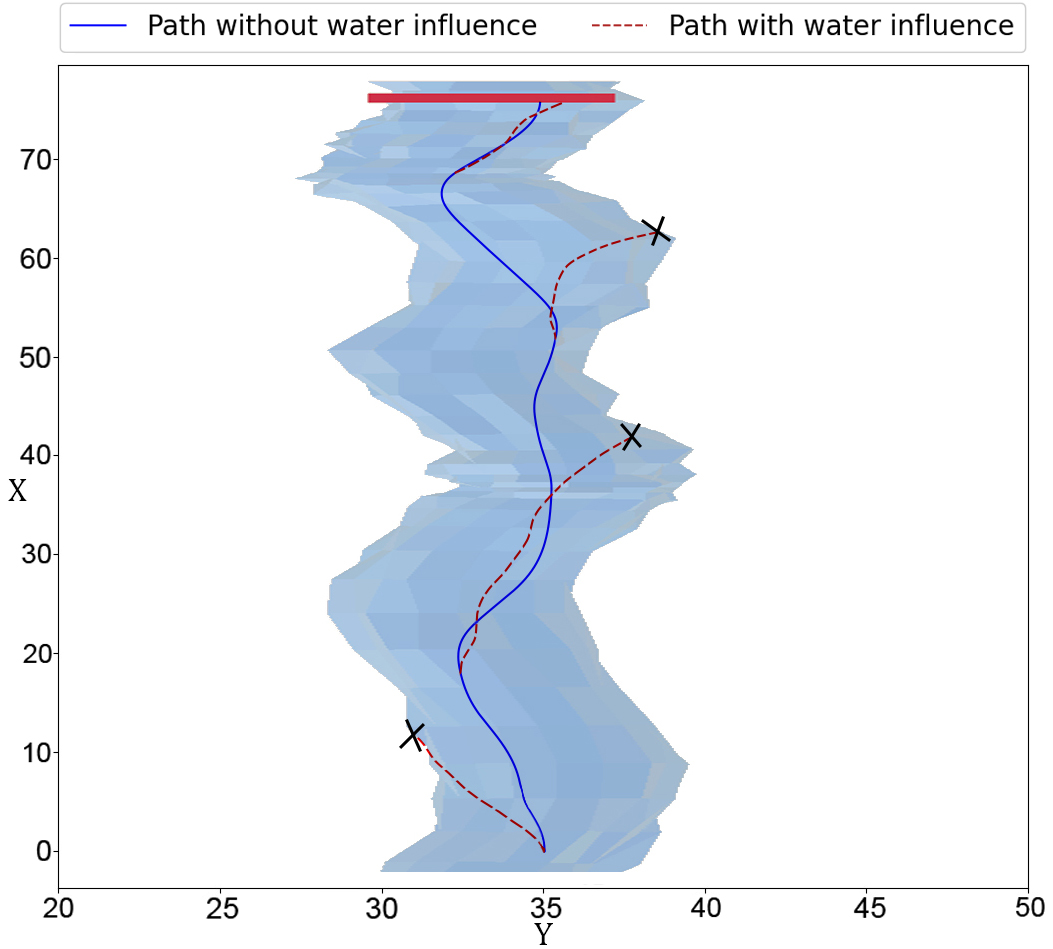}
        \end{center}
    \vspace{-15pt}
    \caption{Illustration of the influence that water exerts on the AUV as it attempts to follow an ideal path.}
    \vspace{-5pt}
    \label{fig:robot:water_impact}
\end{wrapfigure}
Through rigorous experimentation with various settings, we identified an optimal configuration that aligned perfectly with our research objectives, paving the way for the subsequent development of our aquatic environment. 
An illustrative example of Zibra's plugin capabilities can be found in Fig.\ref{fig:robot:100_viscosity} and Fig.\ref{fig:robot:0_viscosity}, where we show the interaction between a simple object and a liquid of varying viscosity. 
The underlying mechanics of ZibraAI's operation involve a novel approach to encoding a 3D object into concise vectors, subsequently decoded by a compact neural network to regenerate the original Signed Distance Field (SDF). This innovative technique finds practical application in gaming physics, particularly in particle simulations. 
The Zibra Liquids Pro plugin represents a collaborative synergy between proprietary physical solvers and machine learning-based neural representations of objects \citep{ZibraHowDoesWorks}. Additionally, ZibraAI has internally developed fluid simulation technology utilizing the Moving Least Squares Material Point Method \citep{hu2018moving}. Early experiments have demonstrated the remarkable efficiency of this approach, capable of simulating 300,000 particles in only 7 milliseconds on a GTX 1050, even without extensive optimizations. The ZibraAI plugin is publicly available, further contributing to the advancement of fluid simulation research, resulting in a critical asset also for robotics and deep learning.

\medskip \textbf{Impact of Water Physics}
We now delve into the influence of the water in our aquatic environment, which represents the critical challenge for our learning agent. In particular, we consider an underwater navigation scenario where a rover must safely navigate in an underwater cave. Understanding this detail is crucial for quantifying the level of unpredictability in the underwater environment and, consequently, assessing the challenges the agent faces in making decisions in this non-stationary scenario. To demonstrate how marine currents can influence AUV trajectories we conducted an additional experiment, illustrated in Fig.\ref{fig:robot:water_impact}. We created an ideal trajectory (blue line) by manually moving an AUV unaffected by water forces. Subsequently, we instructed a second AUV to follow the same sequence of actions as the first, with the additional challenge of marine currents (red dashed line). As shown in the figure, this second rover collided with the cave walls a total of three times. This result underscores the critical importance of generating an intelligent agent capable of dynamically correcting unexpected trajectories that could potentially bring the rover to operate too close to cave walls.

%% file: sections/training.tex
\section{Training Approach}
\label{sec:training}

In this section, we introduce our deep reinforcement learning pipeline, showing the various strategies we employed to develop a safe and reliable agent that serves as a stable baseline for our benchmarking environment. In the following sections, we discuss different approaches describing their respective strengths and weaknesses. Our comparative analysis has been performed to meet the standard requirements for an empirical DRL evaluation \citep{henderson2018deep}; in particular, we report the average reward with the standard deviation from different random initializations for the neural networks (i.e., $10$ different random seeds for each set of experiments). In the following sections, we conduct a comparative analysis focusing solely on the underwater cave exploration sub-domain available in our simulator. The choice is motivated by the fact that underwater navigation is more complex than surface navigation. Indeed, additional factors such as controlling the diving motion and adjusting pressure based on the depths reached need to be considered. These additional difficulties allow us to conduct a more meaningful analysis. Nevertheless, in Sec. \ref{sec:validation} we perform a validation step on the surface navigation problem, confirming our findings.

In both our benchmarks, the goal for the agent is to reach a target destination without colliding with obstacles. Our agent is equipped with 28 sensors arranged in the 180-degree frontal field, allowing it to observe the immediate surrounding environment directly. It is aware of the direction of the target point in a straight line and also knows its own linear and angular velocity. At each time step, the model determines which actions to take by selecting them from a discrete action space, enabling the agent to move forward, rotate, or adjust its depth when submerged. The configuration of sensors and actuators results in a vector observation of $31$ real values; while the action space can be tuned by the user and consists of a variable set of discrete actions.

\subsubsection*{Cave Environments for Training and Testing}
The primary objective of our agent is to navigate through a cave and safely reach the target point without colliding with rocks and walls. Crucially the agent is provided with the coordinates of the destination in terms of polar coordinates to its position; this setup is widely adopted in the literature and constitutes a challenging benchmark for the training agent \citep{RaAcAm19, AmCoYe23, MaFa21}. Achieving this goal demands high capabilities in obstacle avoidance and the ability to counter the unpredictable movements induced by water currents. To comprehensively evaluate our model's performance, we have designed various cave models. Some of these caves serve for the training phase, while others are used for testing purposes. The idea behind this diversity is to expose the agent to a broad spectrum of scenarios, each posing unique challenges. All the training caves exhibit distinct characteristics that serve as robust evaluative metrics for our model. \textit{The first cave} features larger dimensions compared to the others, moreover, it does not present any additional forces due to the currents of the water. The agent can thus focus completely on the simple control aspect. \textit{The second cave} comprises a sequence of narrow passages interspersed with wider areas. This configuration introduces the additional challenge of navigating through tunnels of varying difficulty. \textit{The third cave} presents a long series of curves, each posing different levels of difficulty; crucially, the parameters related to the velocity of the water particles are raised significantly, posing a significant challenge for the agent. Moreover, in addition to the custom caves for training and testing, in Sec.\ref{sec:validation} we perform an additional evaluation using a 3D model built from data from a real cave. This test aims to assess the ability of the agent to navigate safely in complex and realistic environments. All these caves and settings are available for testing in our simulation engine. The set of hyperparameters employed for the training is reported in the public repository\footnote{\url{https://github.com/dadecampo/aquatic_navigation_envs}}; these values have been empirically tuned through a grid search process over a set of common configurations.

\begin{figure}[t]
\minipage{0.3\linewidth}
    \includegraphics[width=\linewidth]{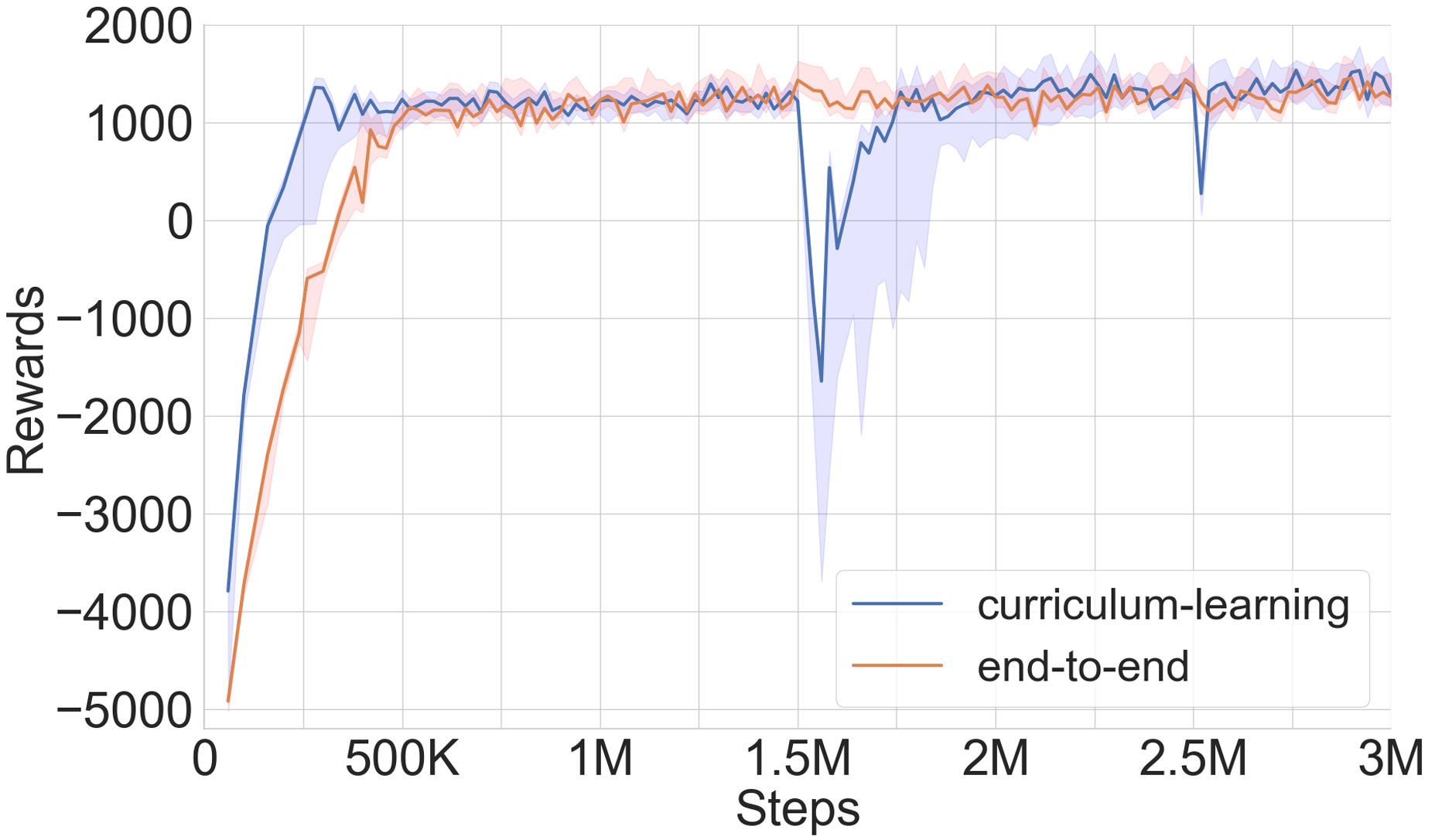}
    \caption{Comparison between curriculum learning and E2E.}
    \label{fig:training:curriculum_results}
\endminipage \hfill
\minipage{0.3\linewidth}
    \includegraphics[width=\linewidth]{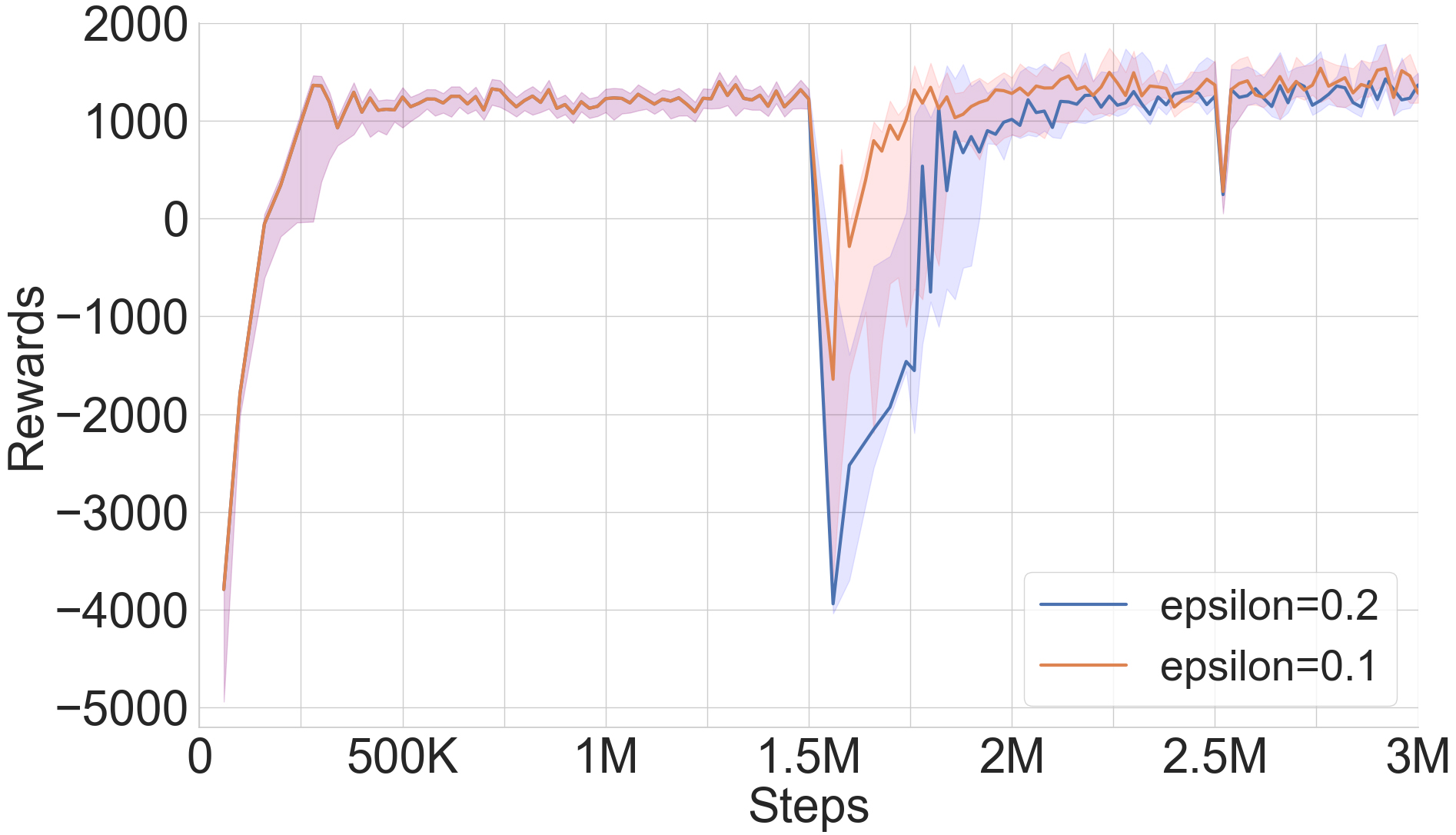}
    \caption{Ablation study on the PPO-clip hyperparameter.}
    \label{fig:training:curriculum_clip}
\endminipage \hfill
\minipage{0.3\linewidth}
    \includegraphics[width=\linewidth]{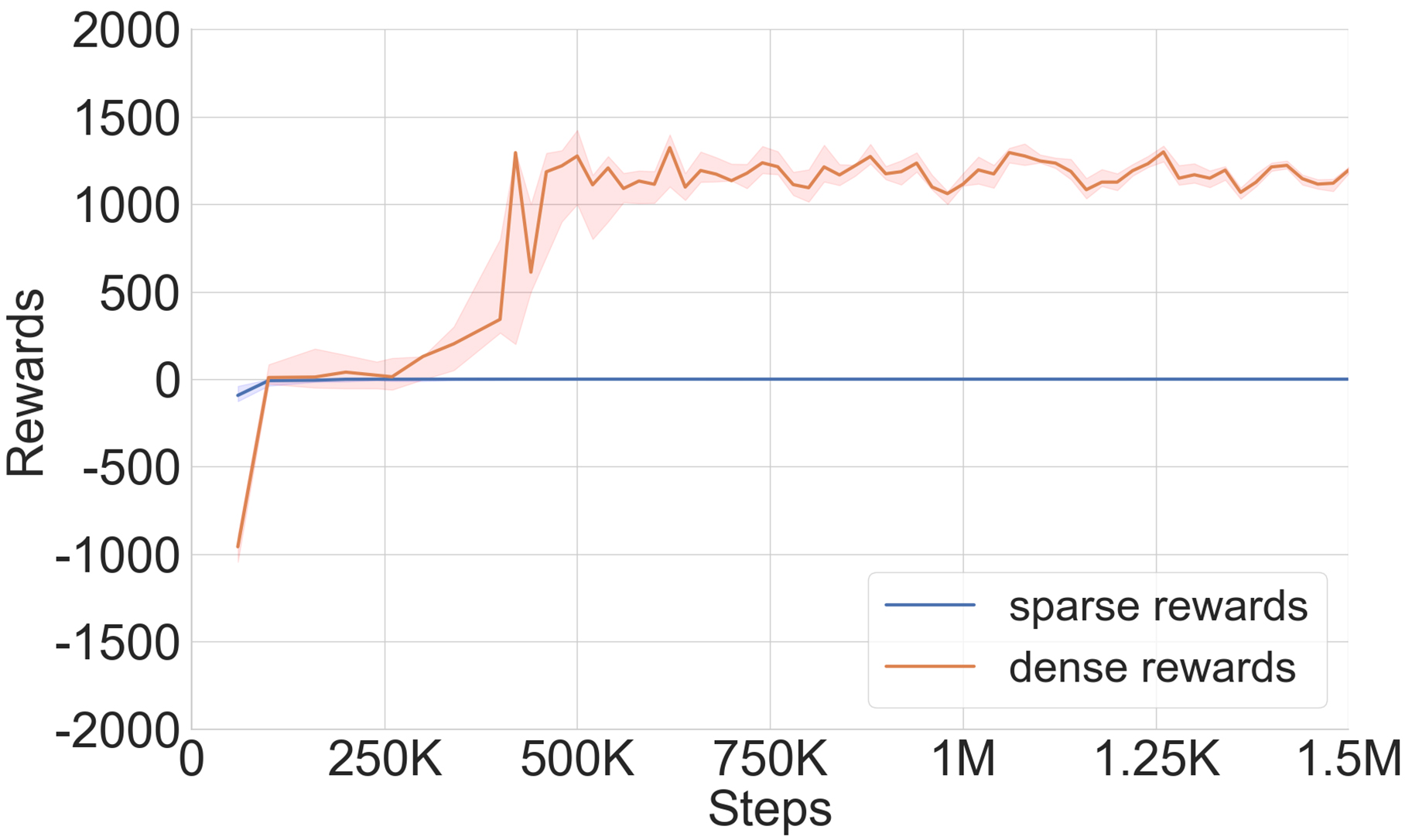}
    \caption{Comparison between sparse and dense reward functions.}
    \label{fig:training:reward_results}
\endminipage
\end{figure}

\subsubsection*{Curriculum Learning} We start by comparing two distinct training approaches: \textit{curriculum learning (CL)} and \textit{end-to-end (E2E)} training. The E2E approach involves training an Artificial Neural Network (ANN) from start to finish on the entire task, without decomposing it into separate subtasks learned sequentially, this leads to simplifying features and reward engineering. Another improvement we made among the different phases of the curriculum regards the PPO-clip value. Specifically, we reduce the clip value from 0.2 (as recommended in the literature) to $0.1$. More details on the approaches and a detailed comparison between the end-to-end approach and our suggested curriculum learning method can be found in Appendix \ref{sec:app:curriculum} of the supplementary materials. To summarize, from our experiments, the curriculum learning approach only slightly reduces the convergence time but it does not provide a substantial improvement in performance (see Fig.\ref{fig:training:curriculum_results}). Interestingly, however, the curriculum learning approach results in a significant improvement from a safety perspective (e.g., the number of collisions with rocks); demonstrating a higher generalization capability in previously unseen environments, as can be seen in Sec.\ref{sec:validation}.

\smallskip \textbf{Conclusion:} For our final experiments we adopted a curriculum learning strategy. According to the literature in the field, our findings suggest that a more structured training pipeline strongly supports a faster and more effective training process \citep{morad2021embodied}.

\subsubsection*{Reward Engineering}
Reward engineering is a pivotal component of a successful deep reinforcement learning (DRL) process. However, the formulation of effective reward functions is often non-trivial, demanding meticulous consideration of various factors. In Appendix \ref{sec:app:reward-engineering} of the supplementary materials we report a detailed comparison between a sparse and a dense reward function, showing the strengths and the weaknesses of both methodologies. To summarize, the training with sparse rewards did not yield success, with the rover notably failing to reach the final goal (see Fig.\ref{fig:training:reward_results}). Conversely, training conducted with dense rewarding has proven to be fruitful, demonstrating the ability to achieve convergence without excessive difficulty.

\smallskip \textbf{Conclusion:} For our final experiments we employed a dense function; formally the reward at time $t$ is calculated as follows:
\begin{equation}
r_t =
\begin{cases}
R_{\text{goalReached}} & \text{if the goal is reached} \\
R_{\text{movement}_t} + R_{\text{timestep}} + R_{\text{collision}} & \text{if a collision occurs} \\
R_{\text{movement}_t} + R_{\text{timestep}} & \text{otherwise}
\end{cases}
\end{equation}
The parameters within the reward function have been defined through empirical testing and can be found in the appendix \ref{sec:app:reward-engineering}.

\subsubsection*{The Safety Aspect}
Although we obtained promising results with the approaches proposed in this section, we focused only on the pure performance of the agent, while in this section, we consider an additional requirement, the overall safety. In particular, we focus on two aspects: (i) the number of collisions with rocks and (ii) the average distance between the agent and the walls of the cave.  More details and results about the safety-oriented reward can be found in Appendix \ref{sec:safety} of the supplementary materials. Our findings demonstrate the positive impact of our explicit safety-centric reward function in terms of reducing the number of collisions and increasing the average safety distance from the cave walls. However, it is worth emphasizing that our approach does not entirely eliminate unsafe behaviors, highlighting the need for future research in this direction.

\smallskip \textbf{Conclusion:} To enhance the agent's safety, we implemented a reward function that considers the distance from the walls as a cost to minimize during training. Crucially, the final training setup used for our real-world evaluations, reported in Sec. \ref{sec:validation}, consists of the previous three training techniques in combination.

%% file: sections/validation.tex
\begin{figure}[b!]
\minipage{0.45\linewidth}
    \includegraphics[width=\linewidth]{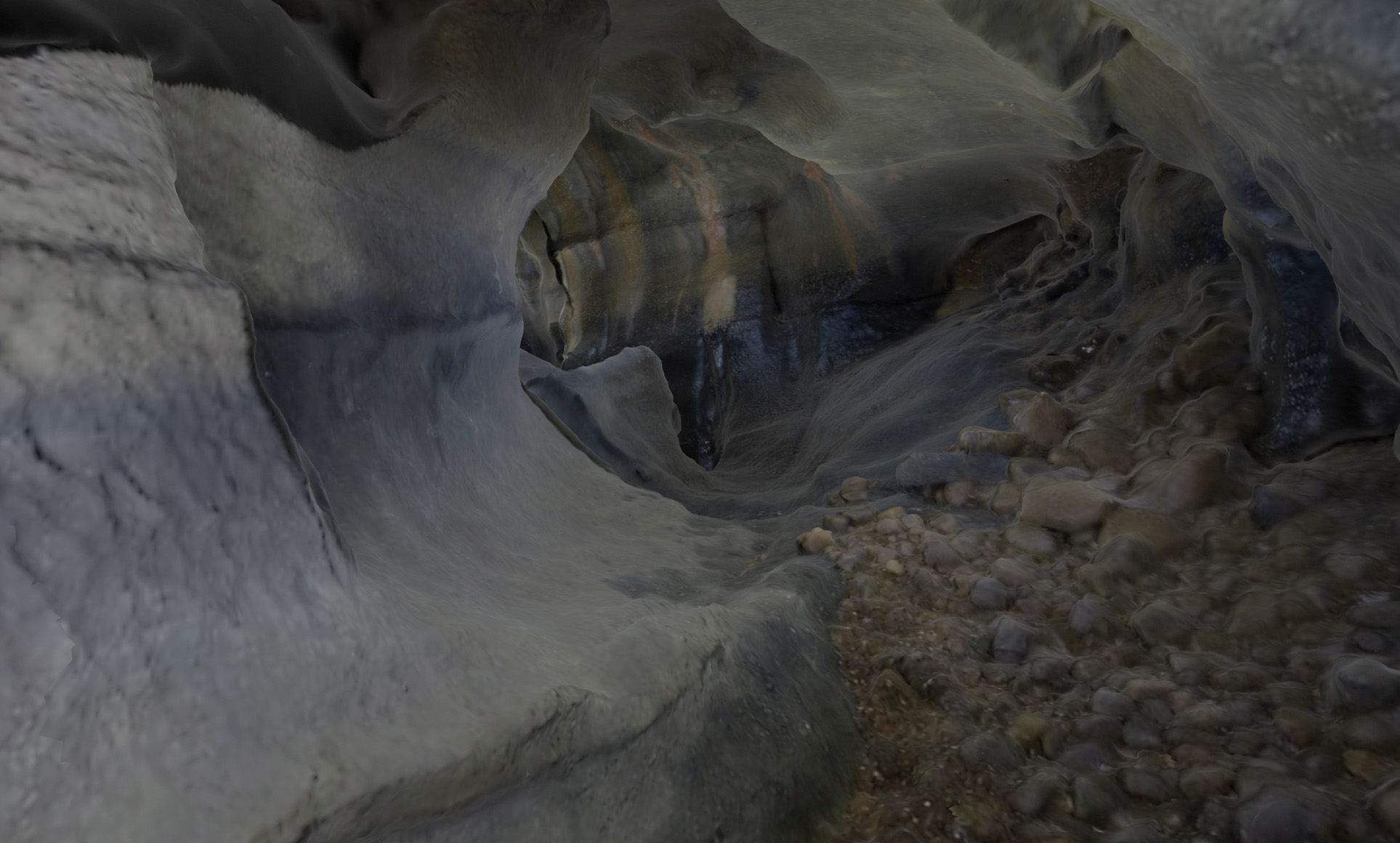}
\endminipage \hspace{1.5cm}
\minipage{0.4\linewidth}
    \includegraphics[width=\linewidth]{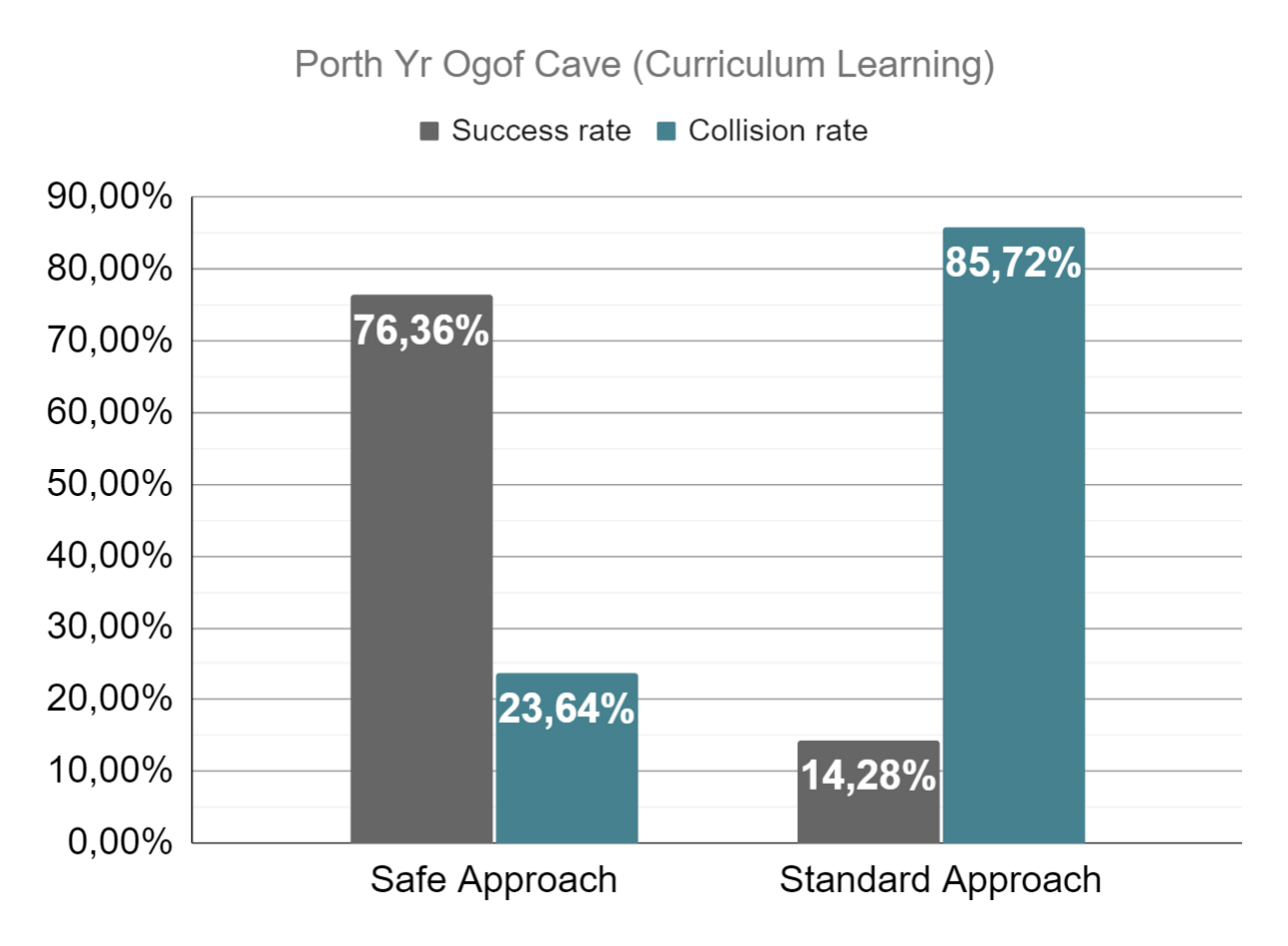}
\endminipage
\caption{On the left is a screenshot of the Porth Yr Ogof cave; on the right are the results obtained by the trained agent.}
\label{fig:validation:cave}
\end{figure}

\section{Evaluation in scenarios built from real-world data}
\label{sec:validation}
In this section, we exploit all the techniques and methodologies discussed throughout this paper to assess the agent's performance within the three-dimensional representation of a real cave. This validation step is made possible through the application of photogrammetry, a technique that enables the creation of highly accurate three-dimensional models using sequences of images or videos. In our simulator, we recreated a detailed portion of Porth Yr Ogof, a cave situated in South Wales \citep{PorthYrOgof}. Porth Yr Ogof is a cave of particular interest due to its unique characteristics. The complex formation of this cave is due to the frequent floods caused by the overflowing of the adjacent Afon Mellte River (the reconstruction of the cave is based on a 3D scan of the area). In this environment, the agent's goal is to reach the end of the cave and thus explore the entire map. What makes the task challenging is that the drone does not have access to the map of the environment, the agent must rely entirely on local observations, exploiting in the decision-making process the policy learned by exploring the caves used for training.

In Fig.\ref{fig:validation:cave} we show a screenshot from our simulator and a plot of the results obtained by our trained agent; for the experimental evaluation, we deploy our agent multiple times starting from a random position of the cave and collecting the average success rate (i.e., the number of time the AUV manages to exit the caves normalized by the number of experiments). We note that, for this evaluation phase, in order to highlight when safety constraints are violated, we considered a single collision as a complete failure.

\begin{figure}[t]
\minipage{0.4\linewidth}
    \includegraphics[width=\linewidth]{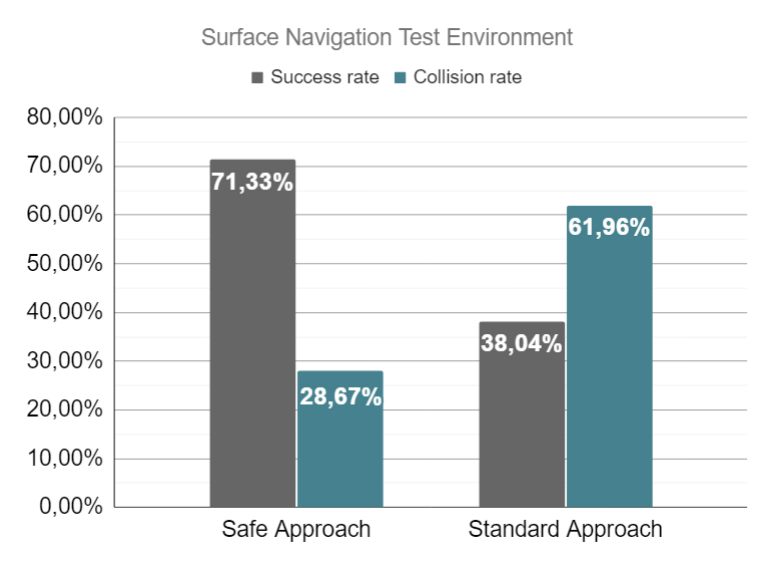}
\endminipage \hspace{1.5cm}
\minipage{0.45\linewidth}
    \includegraphics[width=\linewidth]{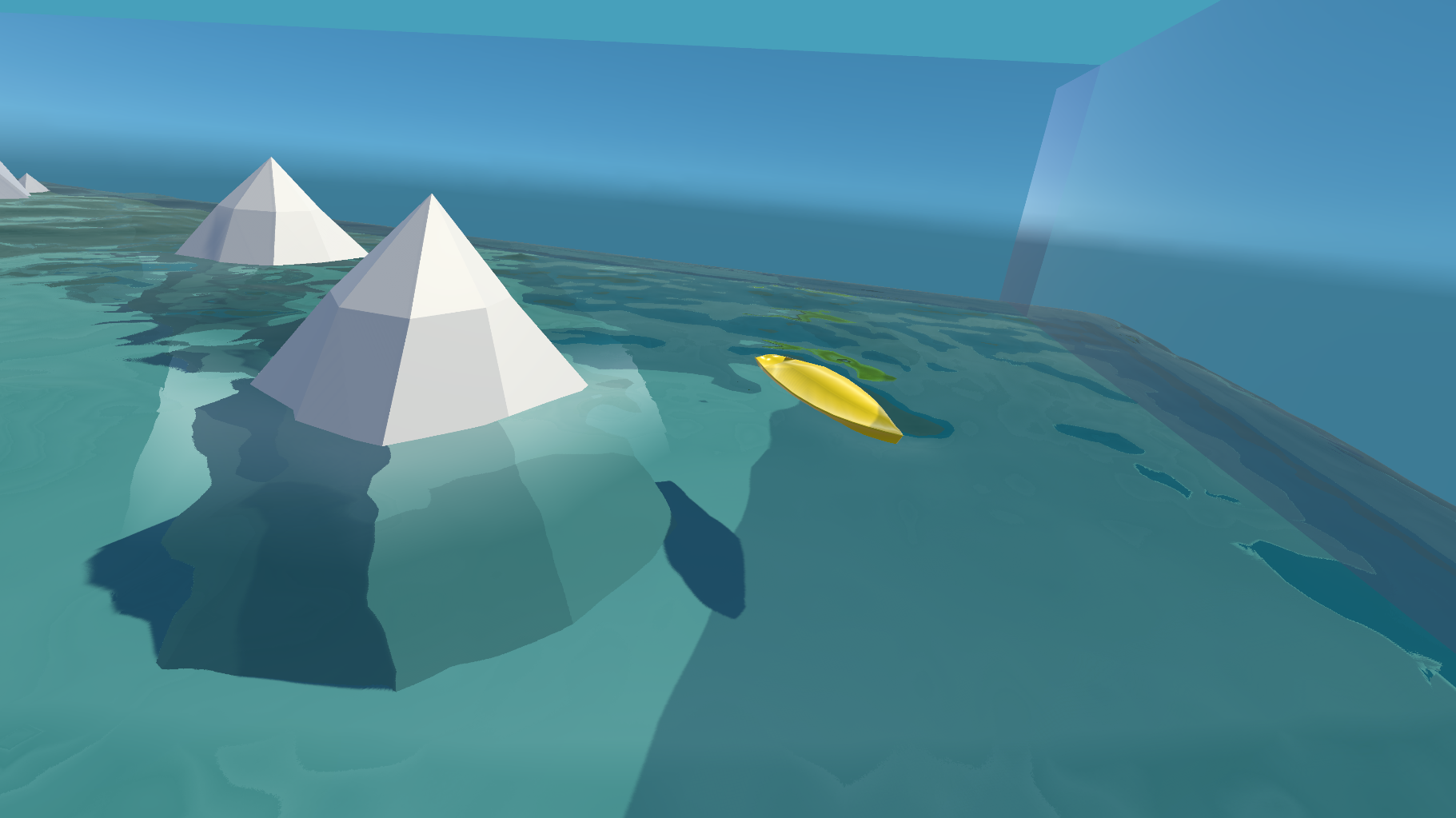}
\endminipage
\caption{On the left are the results obtained by the trained agent; on the right: a screenshot of the surface navigation benchmark.}
\vspace{-5pt}
\label{fig:validation:surface}
\end{figure}

\noindent \textbf{Extention to Surface Navigation} In the last few sections, we have focused on underwater cave exploration because our preliminary experiments have shown it to be the most challenging environment. However, surface navigation presents similar interesting challenges due to the non-stationary and dynamic nature of water. In this last evaluation, we repeat the analysis of the previous problem in this second benchmark. In this environment, the goal of the agent is to reach the target position while avoiding collisions with rocks and reefs. The agent can only rely on observations from local sensors, which include a GPS and compass for computing heading and distance to the target position, and a proximity sensor for detecting obstacles in specific directions. At the initialization of each episode, the map and the positions of the target and agent are randomly generated. Our results are shown in Fig.\ref{fig:validation:surface}, although overall simpler, our results confirm that even in this scenario PPO struggles to find an optimal policy, especially from the safety perspective.

%% file: sections/conclusion.tex
\section{Conclusion}
This paper has presented a challenging benchmark to stimulate the advancement of DRL methods for robot control. Our contributions span three key areas: i) we developed a realistic simulator tailored to the unique challenges of underwater cave exploration and surface navigation; ii) we provided a comprehensive pipeline for training autonomous agents using Deep Reinforcement Learning (DRL); iii) we addressed safety through two critical aspects: collision avoidance and maintaining a safe distance from cave walls. To demonstrate the effectiveness of our approach, we finally conducted an extensive testing phase in a simulation of the real-world cave environment of "Porth yr Ogof" in South Wales where our trained agent successfully explored the cave, avoiding catastrophic collisions with rocks and maintaining a safe distance from cave walls. These contributions together serve to introduce a novel benchmark for deep reinforcement learning in a challenging and realistic scenario and a series of techniques to provide a stable and reproducible result that provides an initial baseline for future development. 

We believe this work paves the way for several future directions, including the exploration of alternative approaches to ensure the safety of our trained agents, such as shielding \citep{AlBlEh18}, constrained reinforcement learning \citep{AcHeTa17}, explainability \citep{bassan2023formally}, and formal verification tools \citep{CountingProVe, CoMaFa21, KaHuIb19, amir2021towards}. Additionally, a natural direction is to move from the simulated environment to real robotic platforms so to gain insights into how the agent interacts with the environment in the real world with the idea of enhancing our simulator. In this last direction, many new challenges arise, such as generalization to unseen situations \citep{amir2023verifying} and possible delays in the communication between the drone and the controller \citep{karamzade2024reinforcement}. Crucially, our simulation tool is freely available for future research and collaborations.

%% file: appendices/curriculum-learning.tex
\begin{figure}[b]
\centering
\includegraphics[width=0.8\linewidth]{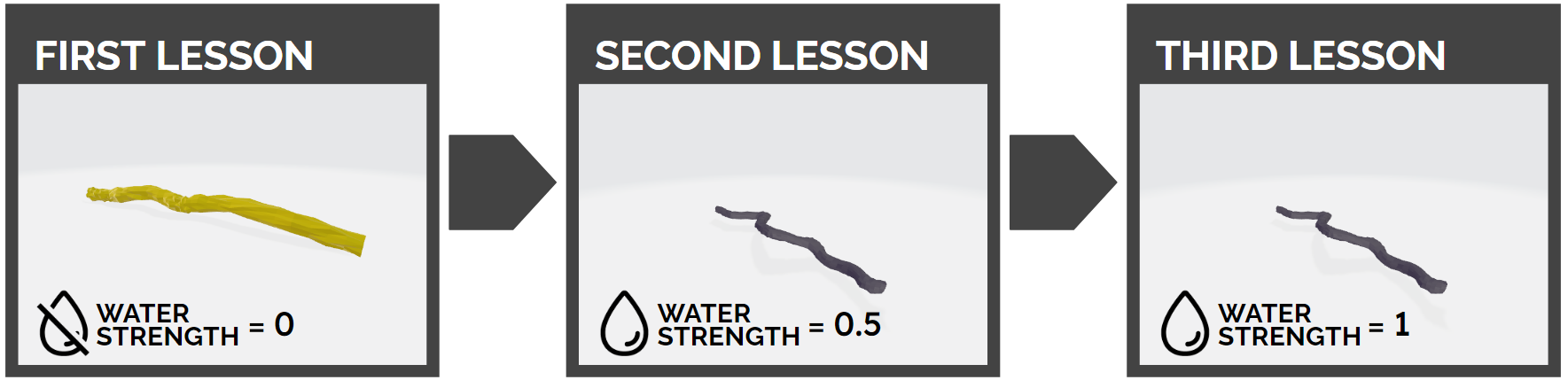}
\caption{Curriculum Learning lessons plan. From left to right the pictures represent the training caves ordered by growing difficult.}
\label{fig:training:curriculum_caves}
\end{figure}

\section{Curriculum Learning}
\label{sec:app:curriculum}
We start by comparing two distinct training approaches: \textit{curriculum learning (CL)} and \textit{end-to-end (E2E)} training. 

The E2E approach involves training an Artificial Neural Network (ANN) from start to finish on the entire task, without decomposing it into separate subtasks learned sequentially. End-to-end training offers the advantage of simplifying feature and reward engineering and reducing the need for manually designing intermediate steps. However, it often necessitates a substantial amount of training data and can pose challenges in terms of interpretability. Curriculum learning, on the other hand, consists of training the agent in a sequence of problems, typically of increasing difficulty; this enables the agent to leverage knowledge and skills acquired in simpler tasks to enhance learning and performance when facing the complete problem. Our curriculum learning pipeline starts with the simplest environment, denoted as \textit{Cave$\_$Train1} (Fig.\ref{fig:training:curriculum_caves}), where the agent learns a set of fundamental skills such as keeping a safe distance from walls and navigating toward the target while executing gentle turns. Subsequently, the weights learned in the initial task are transferred to the new neural network as we progress to the \textit{Cave$\_$Train2} environment. Here, the agent exploits the already learned capabilities in a more intricate environment. Additionally, the rover encounters water currents for the first time, albeit at half the strength intended for the evaluation phase. The third and final phase of our training process introduces the rover to water currents at the intended evaluation strength. This phase can be regarded as a refinement stage, building upon the model developed during the initial two phases. To prevent any loss of knowledge acquired throughout the entire training process, we lowered the learning rate for this final step. 

Another improvement we made among the different phases of the curriculum regards the PPO-clip value. Specifically, we reduce the clip value from 0.2 (as recommended in the literature) to 0.1. The intuition behind this modification is that during the initial \textit{Cave$\_$Train1} phase, the rover learns fundamental movement policies, which we aim to preserve as the training progresses to the more complex stages. By lowering the clip value, we aim to maintain policy stability. However, reducing this parameter requires a tradeoff that lies in a loss in data efficiency (i.e., slowing the acquisition of new behaviors). Typically, knowledge transfer across phases is accompanied by the freezing of layers trained in the previous stage. However, given that the fundamental objective of our agent remains unchanged despite the changing environments, we opted to avoid this technique, allowing for continuous learning and adaptation throughout the curriculum \cite{Goutam2020LayerOutFL}.

\subsubsection*{Results:} Fig.\ref{fig:training:curriculum_results} presents the results of our comparison between the curriculum learning and the E2E approach. Noticeable drops in rewards correspond to the transitions between lessons, occurring at the 1.5 million and 2.5 million timestep marks. However, in both cases, the reward graphs quickly recover and converge to approximately 1200. To ensure a fair comparison, we assign to E2E training the cumulative number of timesteps across all phases of curriculum learning required to achieve a satisfactory result in the analyzed environment. 

The curriculum learning approach slightly reduces the convergence time required but it does not provide a substantial improvement in performance. Interestingly, however, the curriculum learning approach results in a significant improvement from a safety perspective (e.g., number of collisions with rocks); demonstrating a higher generalization capability (Sec.\ref{sec:safety} provides more detail about this result). Moreover, Fig.\ref{fig:training:curriculum_clip} shows an ablation study to motivate our choice of using a smaller PPO-clip value with respect to the standard setting; our results clearly show the obtained improvements, confirming our design choice. 

%% file: appendices/safety.tex
\begin{figure}[b!]
\centering
\includegraphics[width=0.5\linewidth]{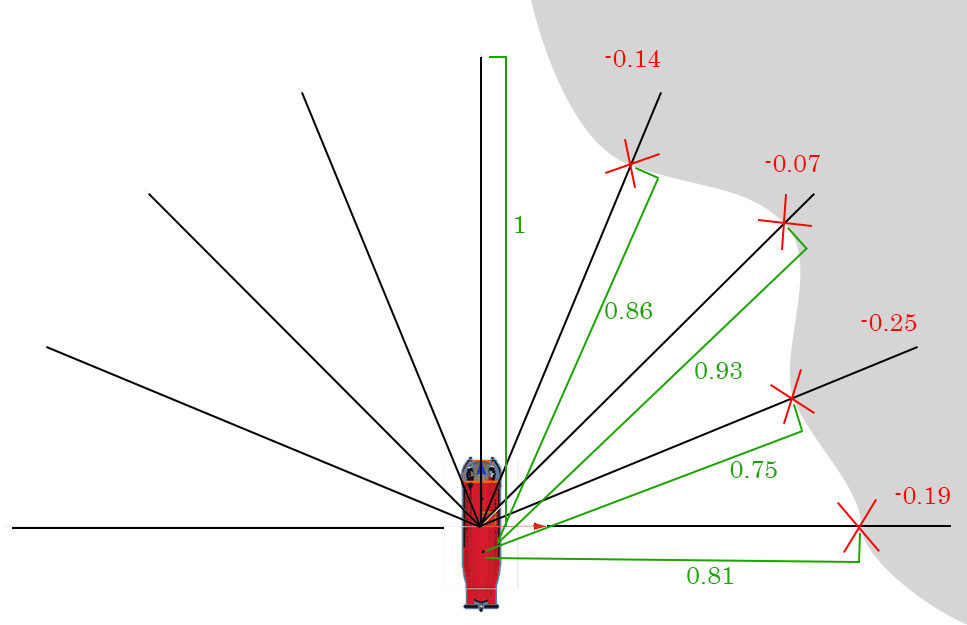}
\caption{This image illustrates the concept of ``interceptionValue", highlighted in green, while the penalties imposed on the agent are indicated in red.
}
\label{fig:safety:lidar-scan}
\end{figure}

\begin{figure}[t!]
\minipage{0.60\linewidth}
    \includegraphics[width=\linewidth]{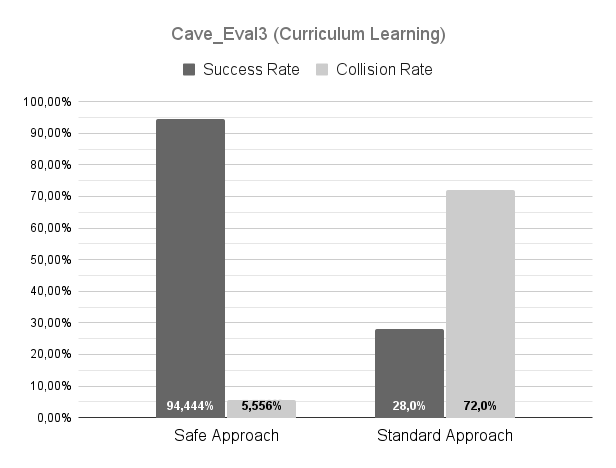}
\endminipage \hfill
\minipage{0.35\linewidth}
    \includegraphics[width=\linewidth]{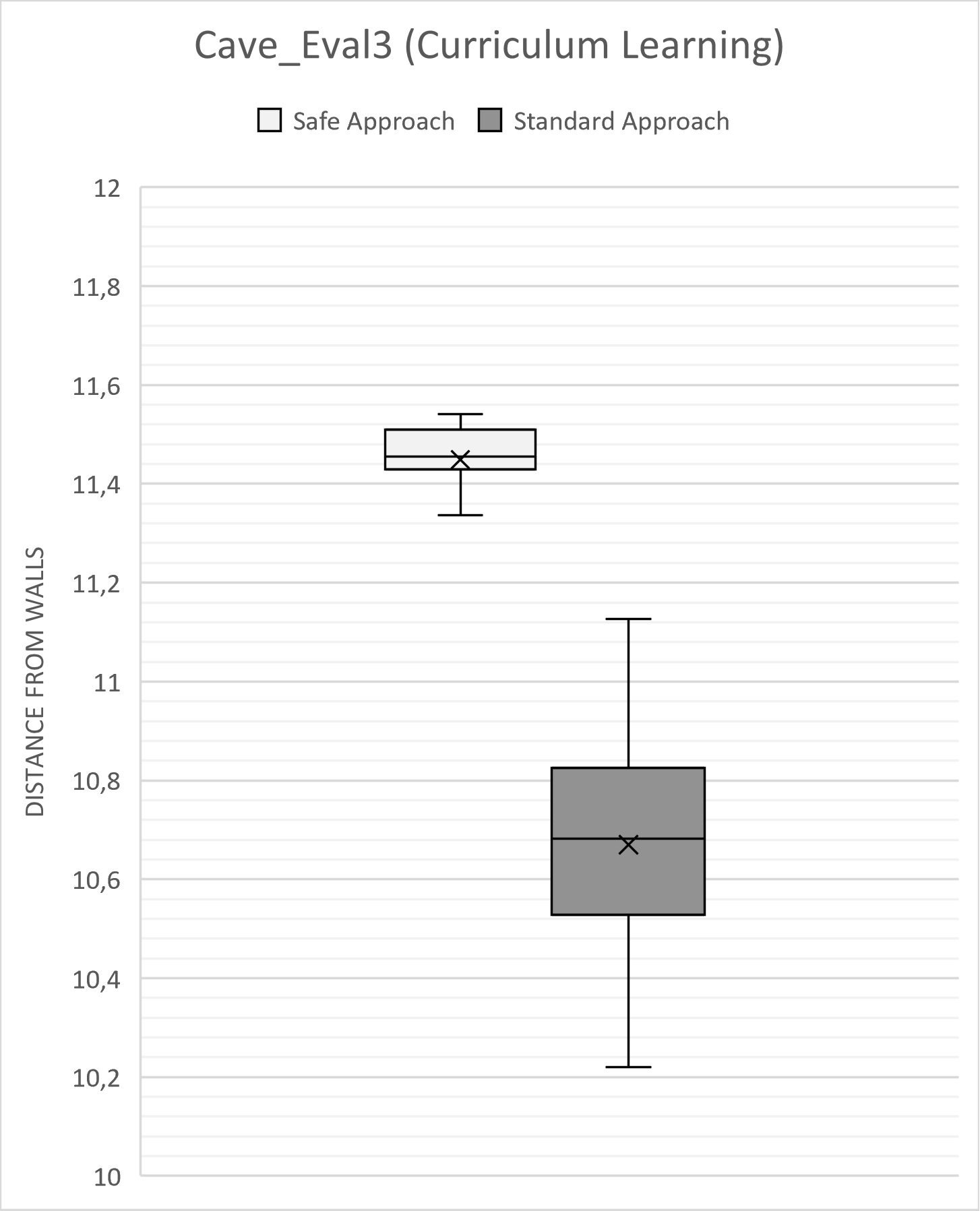}
\endminipage
\caption{Analysis of the agent's performance from a safety perspective.}
\label{fig:safety:results}
\end{figure}

\section{The Safety Aspect}
\label{sec:safety}

Although we obtained promising results with the approaches proposed in Sec.~\ref{sec:training}, we focused only on the pure performance of the agent, while in this section, we consider an additional requirement, the overall safety. In particular, we focus on two aspects: (i) the number of collisions with rocks and (ii) the average safe distance between the agent and the walls of the cave. In the literature, numerous approaches exist to improve the safety of a learning agent, such as Constrained Deep Reinforcement Learning \cite{StAcAb20}, Safe Exploration \cite{SiJaSp21}, Shielding \cite{AlBlEh18}, and Formal Verification \cite{KaHuIb19}; however, these techniques go beyond the benchmarking scope of this work, and we leave the analysis of these approaches for future research. In contrast, in this paper, we focus on the concept of ``reward engineering". In particular, we propose to modify the reward function presented in Sec.~\ref{sec:training}, refining it to encourage more cautious behaviors. Through this rewarding mechanism, we emphasize the role of the proximity sensors, treating them not only as part of the observation space but also as key components in the calculation of the reward function. By doing so, we expect our agent to exhibit safer behavior, actively attempting to maintain a safe distance from the cave walls.

\[
r_t =
\begin{cases}
R_{\text{goalReached}} & \text{goal} \\
R_{\text{movement}_t} + R_{\text{timestep}} + R_{\text{collision}} + R_{\text{sensors}} & \text{collision} \\
R_{\text{movement}_t} + R_{\text{timestep}} + R_{\text{sensors}}  & \text{otherwise}
\end{cases}
\]

The term $R_{\text{sensors}}$ is calculated based on the measurements taken at each timestep by the rover's sensors. When contact occurs through the lidar sensor, a value is returned indicating the height at which the ray was intercepted. This value is normalized to 1, and then, by using $-(1 - \text{rayInterceptionValue})$, subsequently multiplied by a constant. This operation is repeated for each of the 28 rays at every single timestep. The way this reward has been adjusted, based on the multiplication by the chosen constants, limits the range of $R_{\text{sensors}}$ to $(-0.6, 0.0]$. Figure~\ref{fig:safety:lidar-scan} provides a visual explanation of the lidar sensor readings of our robot.

\subsubsection*{Results:} In this section, we delve into the results of our safety-oriented analysis by focusing on the results obtained in the most challenging scenario, referred to as \textit{``cave\_test3"}. Results are presented in Figure~\ref{fig:safety:results}. Our findings demonstrate the positive impact of our safety-centric reward function in terms of reducing the number of collisions and increasing the average safety distance from the cave walls. However, it is worth emphasizing that our approach does not entirely eliminate unsafe behaviors, highlighting the need for future research in this direction.

%% file: appendices/reward-engineering.tex
\section{Reward Engineering}
\label{sec:app:reward-engineering}

Reward engineering is a pivotal component of a successful deep reinforcement learning (DRL) process. However, the formulation of effective reward functions is often non-trivial, demanding meticulous consideration of various factors. For example, an important challenge revolves around achieving the delicate balance between shaping the agent's behavior and avoiding inadvertent side effects. An excessively simplistic or sparse reward may prevent an effective learning process. Conversely, an overly intricate function may result in an agent incapable of generalizing beyond the training environment \cite{hu2020learning}. Another well-known problem pertains to reward hacking, i.e., whereby an agent exploits loopholes or biases in the reward function to maximize rewards without genuinely fulfilling the intended task. This underscores the imperative need for crafting reward functions that are both informative and resilient to manipulation. In literature, reward functions are typically subdivided into two main categories: sparse and dense. In this section, we conduct an analysis comparing the efficacy of these two reward paradigms applied to our case study.

\subsubsection*{Sparse Rewards}
Sparse rewards consist of a structure where the agent receives a reward signal only upon accomplishing specific operations (e.g., avoiding an obstacle) or achieving particular objectives (e.g., reaching a target position). However, for the majority of the learning process, the agent encounters limited to no feedback. Sparse rewards present notable challenges for RL agents due to their nature, offering minimal guidance during the learning process. For example, an agent may struggle to understand which actions or states contribute to their success or failure, given the sporadic feedback. Consequently, this can lead to a slow learning loop or hindered convergence, particularly in complex environments. Despite these limitations, sparse rewards are easy to design and are more robust against reward hacking.

We now introduce the discrete reward function we employed in our training process. This is a straightforward reward system that grants rewards to the rover under specific conditions while imposing penalties for undesirable outcomes (e.g., collisions with rocks or cave walls).
\begin{equation}
r_t =
\begin{cases}
R_{\text{goalReached}} & \text{if the goal is reached} \\
R_{\text{collision}} & \text{if the rover collides} \\
R_{\text{fail}} & \text{timeout}
\end{cases}
\end{equation}
In this reward function, $R_{\text{goalReached}}$ represents the reward granted when the rover successfully reaches its goal, $R_{\text{collision}}$ signifies the penalty for collisions, and $R_{\text{fail}}$ denotes the penalty imposed if the rover fails to achieve the goal within the stipulated time frame. In our setup, we impose $R_{\text{goalReached}}=10$, $R_{\text{collision}}=-10$, and $R_{\text{fail}}=-1$.

\begin{figure}[tb]
\minipage{0.58\linewidth}
    \includegraphics[width=0.6\linewidth]{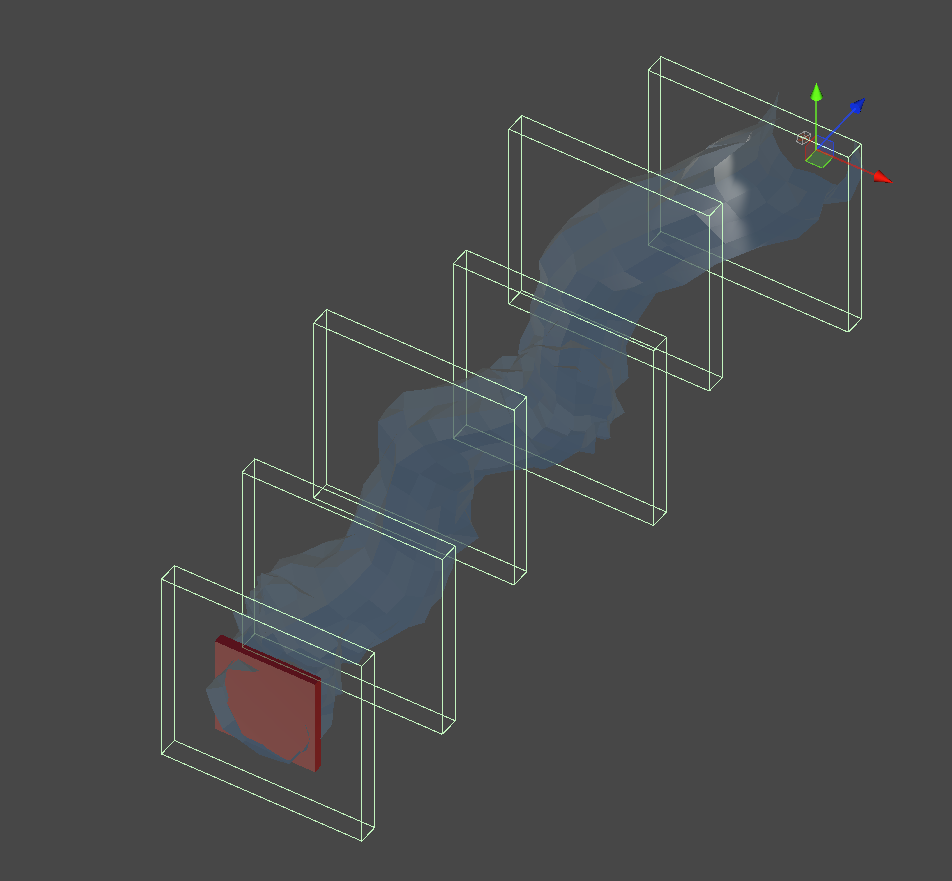}
\endminipage \hfill
\minipage{0.33\linewidth}
    \includegraphics[width=0.6\linewidth]{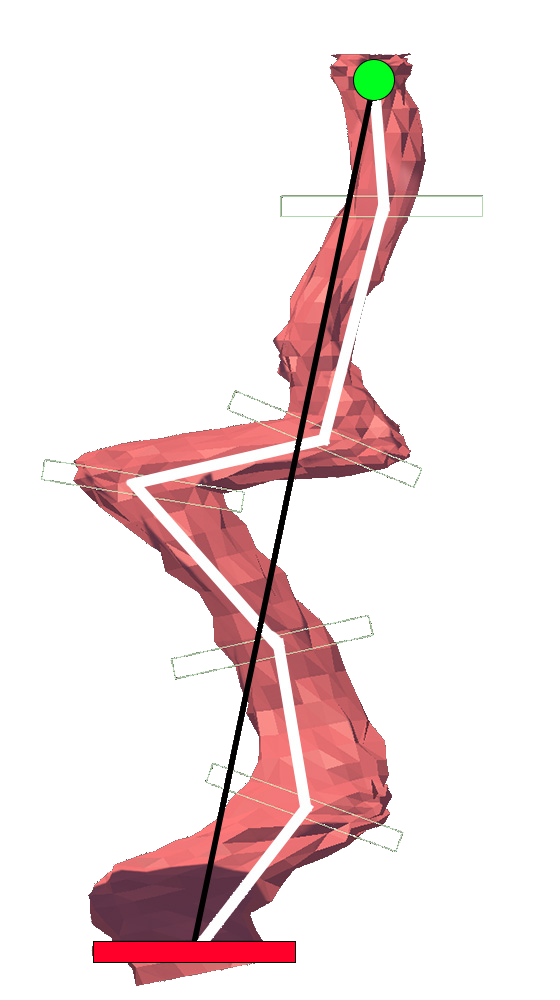}
\endminipage
\caption{Comparison between \textit{DistanceRewarder} (white) and Euclidean distance (black line).}
\label{fig:training:reward_distance_comparison}
\end{figure}

\subsubsection*{Dense Rewards}
Dense rewards represent an alternative structure where the agent receives a signal more frequently, typically at each time step of an episode. These rewards exhibit a higher level of continuity and furnish constant feedback to the agent as it progresses through the environment. We now introduce the continuous function we proposed in our work. This reward incentivizes movement toward the cave exit while penalizing proximity to cave walls. During training, an accurate calculation of the distance from the goal is crucial, and to achieve this, we introduce a system of panels that trace the tunnel's curvature (Fig. \ref{fig:training:reward_distance_comparison}). By measuring the distance between these panels, along with the distance from the last panel to the rover, we obtain a more precise approximation of the distance from the goal. This system, termed ``DistanceRewarder", is a specific and fundamental component we developed to furnish a more precise reward to the agent. 

Moreover, the reward function we employed for our standard training includes multiple additional components designed to incentivize specific behaviors. First, a specific signal that encourages the rover to progress along the path toward the goal. At each timestep, the agent is rewarded with a value proportional to the distance traveled toward the cave exit. It's important to note that moving away from the goal results in a penalty. Second, to promote behavior that takes the environment into account, a penalty is applied when the agent collides with a cave wall. The third component of the function is a reward bonus, which is obtained upon reaching the cave exit. During training, the reward at time $t$ is calculated as follows:

\begin{equation}
r_t =
\begin{cases}
R_{\text{goalReached}} & \text{if the goal is reached} \\
R_{\text{movement}_t} + R_{\text{timestep}} + R_{\text{collision}} & \text{if a collision occurs} \\
R_{\text{movement}_t} + R_{\text{timestep}} & \text{otherwise}
\end{cases}
\end{equation}

For our experiment we set $R_{goalReached}=500$, $R_{movement_t}=(distanceToGoal_{t-1}-distanceToGoal_{t})*10$, $R_{timestep}=-0.01$ and $R_{collision}=-0.01$. $R_{movement_t}$ rewards the displacement of the rover in the direction of the goal. In our environment, the combination of rover speed and water friction puts this reward in a range from -0.03 to 0.03, so we can conclude $R_{movement_t} \in \left(-0.03, 0.03\right)$.

\subsubsection*{Results:} 
We now provide a comparative analysis of dense and sparse rewarding approaches. As depicted in Fig.\ref{fig:training:reward_results}, training with sparse rewards did not yield success, with the rover notably failing to reach the final goal. Conversely, training conducted with dense rewarding has proven to be fruitful, demonstrating the ability to achieve convergence without excessive difficulty. In the next section, we will focus on the safety aspects, showing how the reward engineering process is crucial also for this latter aspect.